\documentclass[10pt, a4paper]{article}

\usepackage[final]{lrec-coling2024} 

\usepackage{makecell}
\usepackage{xcolor}
\usepackage{todonotes}
\usepackage{algorithm}
\usepackage{algpseudocode}
\usepackage[fleqn]{amsmath}
\usepackage{mathabx}
\usepackage{xstring}
\usepackage{color}
\usepackage{graphicx}
\usepackage{tabularx}
\usepackage{booktabs}

\newcommand{\customfont}{\fontsize{8.8pt}{9.5pt}\selectfont}

\title{Gaining More Insight into Neural Semantic Parsing with Challenging Benchmarks}

\name{Xiao Zhang, Chunliu Wang, Rik van Noord, Johan Bos} 

\address{Center for Language and Cognition, University of Groningen \\
         \{xiao.zhang, chunliu.wang, r.i.k.van.noord, johan.bos\}@rug.nl\\}

\abstract{
The Parallel Meaning Bank (PMB) serves as a corpus for semantic processing with a focus on semantic parsing and text generation. 
Currently, we witness an excellent performance of neural parsers and generators on the PMB. This might suggest that such semantic processing tasks have by and large been solved. 
We argue that this is not the case and that performance scores from the past on the PMB are inflated by non-optimal data splits and test sets that are too easy.
In response, we introduce several changes. 
First, instead of the prior random split, we propose a more systematic splitting approach to improve the reliability of the standard test data. Second, except for the standard test set, we also propose two challenge sets: one with longer texts including discourse structure, and one that addresses compositional generalization.
We evaluate five neural models for semantic parsing and meaning-to-text generation. Our results show that model performance declines (in some cases dramatically) on the challenge sets, revealing the limitations of neural models when confronting such challenges.
\\[3pt]
\Keywords{Annotated Corpus, Discourse Representation Theory, Semantic Parsing, Text Generation}}

\begin{document}

\maketitleabstract

\section{Introduction\label{sec: introduction}}

The Parallel Meaning Bank (PMB, \citealp{abzianidze-etal-2017-parallel}) is a semantically annotated parallel corpus for multiple languages. It consists of a large collection of parallel texts, each accompanied by a formal meaning representation based on a variation of Discourse Representation Theory (DRT, \citealp{kamp1993discourse}), called Discourse Representation Structure (DRS). It can be used for corpus-based studies on formal semantic phenomena, or to develop and evaluate semantic processing tasks such as text-to-meaning parsing and meaning-to-text generation.
As a matter of fact, the PMB has been widely used in semantic parsing \citep{abzianidze-etal-2019-first, van-noord-2019-neural, van-noord-etal-2020-character, wang-etal-2021-input, poelman-etal-2022-transparent}, natural language generation \citep{wang-etal-2021-evaluating, wang-etal-2023-pre}, and semantic tagging \citep{bjerva-etal-2016-semantic,abzianidze-bos-2017-towards,abdou-etal-2018-learn,huo-de-melo-2020-inducing}.

The rapid development of neural models and their incredible performance seem to make the impression that tasks like semantic parsing are practically solved. 
For instance, the state-of-the-art DRS parser \citep{wang-etal-2023-pre} achieves a remarkable score of approximately 95.0 on the English test set of the PMB and manual analysis reveals that the parser made very few errors except for words outside the vocabulary. 
Are neural models mastering semantic parsing (and indeed natural language generation), even for complex formal meaning representations like those present in the PMB? Or is there something else going on, and does this perception not align with the actual state of affairs?

\begin{figure*}[hbtp]
    \includegraphics[width=\textwidth]{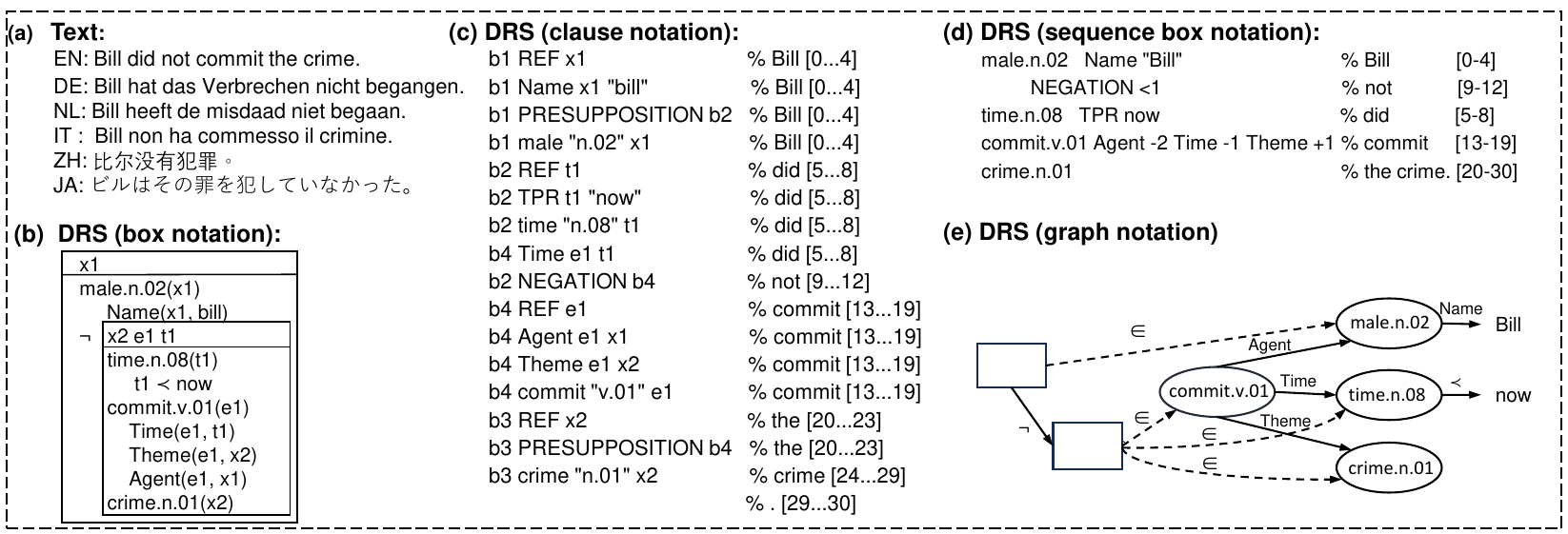} \vspace*{-22pt}
    \caption{(a) An example sentence ``\emph{Bill did not commit the crime.}" taken from the PMB in six languages with its DRS in (b) box notation, (c) clause notation, (d) sequence box notation, and (e) graph notation.}
    \label{fig:PMB-data-example} 
\end{figure*}

We carried out a critical examination of the PMB and revealed three (related) problems: (1) there is a ``data leakage'' from the training data to the development and test splits; (2) the random splits of the data lead to a non-optimal division; and (3) the test set is often regarded as ``easy'' as it contains a large amount of relatively short sentences. Let us elaborate on this a bit.

In the current release of the PMB, the data splits were randomly decided and considered "standard". However, this random split may result in overlap and imprecise error estimates \citep{sogaard-etal-2021-need} and and cannot adequately represent the distribution of the dataset. For instance, the sentence ``\emph{I like chocolate ice cream!}'' is allocated to the training set, while the very similar sentence ``\emph{I like chocolate ice cream.}'' is assigned to the test set. Equally alarmingly, some instances in the development and test sets mirror those in the training set, potentially skewing parser evaluations.  Consequently, this may lead to parser evaluation results that are overly optimistic. 
We completely agree with \citet{opitz-frank-2022-better} and \citet{groschwitz-etal-2023-amr}, who both argue that "AMR Parsing is far from solved" hits the nail on the head, and even goes beyond Abstract Meaning Representation (AMR) and also applies to DRS.
We think the current PMB test set lacks difficulty, because it puts emphasis on brief and simplistic sentences with an average length of less than ten words.
The reason for this is that all instances of the test set have the ``gold'' annotation status, obtained via intensive manual correction, and  the longer a document the harder it is to get an error-free annotation for it.

The aim of this paper is (a) to show that the random split indeed leads to an undesired simplification of the task, and 
(b) to demonstrate that the task of semantic parsing is far from being solved by providing a new challenging test set.

Inspired by the work of \citet{sogaard-etal-2021-need}, we design three new test sets: one standard test set and two challenge sets. 
The former is implemented by a two-round sorting approach to establish a more systematic split, ensuring the reliability and independence of standard development and test sets. The latter comprises a test set with substantially \emph{longer texts} and a test set based on \emph{compositional recombination}. The long-text set is derived by choosing documents with long texts from the PMB and  manually correct the automatically assigned meaning representation.  This set aims to assess the parser's performance on long and multi-sentence texts. The compositional set consists of texts formed by recombining the Combinatory Categorical Grammar (CCG, \citealp{Steedman1996SurfaceSA}) derivation tree that is provided with the PMB data. This kind of tree recombination technique has been empirically validated for semantic data augmentation by \citet{juvekar2023semantically}. Differently, we employ this technology for the creation of test sets, with the intent of assessing the semantic parser model's capability in compositional generalization \citep{Furrer-2020-com}. To our knowledge, we are the first to utilize CCG to create data for compositional generalization  testing. 
By empirical analysis of the performance of neural semantic parsers and generators based on five different language models, we show the effect of our newly created systematic split and challenge sets.

\section{Background and Related Work\label{sec: background}}

In this section, we first provide an overview of DRS, PMB, and CCG, review the works in parsing and generation, and introduce different data split methods. Subsequently, we introduce existing tasks and corpora related to long text semantic and compositional generalization.

\subsection{Discourse Representation Structure\label{subsec:drs}}

DRS is the formal meaning representation in the PMB, capturing the essence of the text and covering linguistic phenomena like anaphors and temporal expressions. Unlike many other formalisms such as Abstract Meaning Representation (AMR, \citealp{banarescu-etal-2013-abstract}) used for large-scale semantic annotation efforts, DRS covers logical negation, quantification, and discourse relations, has complete word sense disambiguation, and offers a language-neutral meaning representation. 


DRS can be represented in multiple formats as is shown in Figure\ref{fig:PMB-data-example}. In the box notation, DRS uses boxes containing discourse referents and conditions. Discourse referents, like \emph{x1}, serve as markers for entities introduced in the discourse. Conditions convey information over the referents: to what concepts they belong and what relations they have to other referents, expressed by roles or comparison operators. Concepts are grounded by WordNet synsets, such as \emph{male.n.02}. Thematic roles are derived from VerbNet \citep{Bonial-2011-verbnet}, for instance \emph{Agent}. Operators, like $\prec$, $=$, $\neq$ and $\sim$, are utilized to formulate comparisons among entities. Furthermore, conditions can also be complex, serving to represent logical (negation, $\lnot$) or rhetorical relations among different sets of conditions.

The clause notation is converted from box notation to adapt to machine learning models \citep{rik-2018-DRS}. In the conversion, the label of the box, wherein the discourse referents and conditions are located, is positioned to precede them.

To simplify DRS, \citet{Bos2023IWCS} introduced a variable-free DRS format called Sequence Box Notation (SBN), where the sequencing of terms is important. The meaning of each word adheres to an entity-role-index structure, with indices connecting entities and roles decorating connection. The discourse relations (such as NEGATION and ELABORATION) are slightly different, indicating the beginning of a new context. The subsequent indices, marked with comparison symbols ($<$,$>$), link the newly established context to another context. SBN can also be interpreted as a directed acyclic graph, as depicted in Figure \ref{fig:PMB-data-example}(e).

\subsection{Combinatory Categorical Grammar}

CCG is a lexicalised grammar formalism \cite{Steedman1996SurfaceSA} used in the PMB to steer the compositional semantics. It comprises just a few basic categories --- N (noun), NP (noun phrase), PP (prepositional phrases) and S (sentence) --- from which function categories can be composed using the backward slash for combining with phrases to the left and the forward slash for combining with phrases to its right. For instance, a typical determiner gets the lexical category NP/N to look for a noun (N) on its right resulting in a noun phrase (NP). CCG expressions can be combined with each other obeying the combinatorial rules, of which there are just a handful. The most common rules are forward and backward application: 
\begin{align}
&\text{Forward App. \quad(>)}: \quad (X/Y) \ Y \Rightarrow X \label{eq: forward}\\
&\text{Backward App. \ (<)}: \quad Y \ (X \backslash  Y) \Rightarrow X \label{eq: backward}
\end{align}

In the PMB, each CCG category is paired with a meaning representation with a semantic type that mirrors the internal structure of the category. This makes it a formidable linguistic formalism to implement compositional semantics.

\subsection{The Parallel Meaning Bank\label{subsec: pmb}}

The PMB has evolved through four versions. Originating from the English-specific Groningen Meaning Bank (GMB, \citealp{basile-etal-2012-platform}), the PMB expanded it by embracing multiple languages. The initial version introduced German, Dutch, and Italian with their gold standard DRS in box format. The second version added silver and bronze standard data, which are partially corrected and uncorrected. Subsequent versions, namely the third and fourth versions, have witnessed an increased volume of manually annotated data and a shift from box to clause notation.

The PMB employs seven layers to process raw text, with each layer contributing an additional piece of syntactic/semantic information, building upon the results from the preceding layer \citep{Abzianidze-2020-pmb}. The seven layers encompass tokenization, symbolization, word sense disambiguation, co-reference resolution, thematic role labeling, syntactic analysis and semantic tagging. Manual corrections are allowed at every layer. The final layer yields a CCG derivation tree, which is then utilized as input for the Boxer \citep{Bos2015OpenDomainSP} and is converted into DRS. Initially tailored for English, PMB aligns it with other languages using an annotation projection method \citep{Abzianidze-2020-pmb}.

In the field of semantic-related tasks, PMB has been widely used. However, it is not without limitations. \citet{haug-etal-2023-long} emphasizes that a large portion of PMB data consists of short sentences, which compromises its ability to accurately represent real-world data.

\subsection{Parsing and Generation with DRS}

Semantic parsing with DRS initially employed rule-based parsers, such as Boxer \citep{bos-2008-wide}. With the advent of neural models, the focus shifted to seq2seq approaches using LSTMs \citep{van-noord-etal-2019-linguistic,van-noord-etal-2020-character}. However, recent innovations include tree-based \citep{liu-etal-2018-discourse, liu-etal-2019-discourse, poelman-etal-2022-transparent} and graph-based techniques \citep{fancellu-etal-2019-semantic, fu-etal-2020-drts}. In the ongoing exploration of neural networks, parsers have increasingly embraced transformer-based models like T5 \citep{Colin-etal-2019-T5}, BART \citep{lewis-etal-2020-bart}, and their variants. A significant breakthrough was DRS-MLM \citep{wang-etal-2023-pre}, a model that pre-trained mBART on PMB data and achieved state-of-the-art results in multiple languages. For meaning-to-text generation, \citet{wang-etal-2021-evaluating} utilized a bi-LSTM on DRS's linearized format and found character-level decoders optimal. The mentioned DRS-MLM can also be used for DRS-to-text generation in pre-training steps outperforming other generators. 

\subsection{Data Split Methods}
In most of the standardized datasets \citep{marcus-1994-penn, fares-etal-2018-ud}, a consistent test set is typically maintained to enable comparisons between models \citep{van-der-goot-2021-need}. Traditionally, this kind of test set is created by random sampling \citep{elazar-goldberg-2018-adversarial, poerner-etal-2018-evaluating}, as is the current practice in the PMB. However, as we mentioned in the introduction, this random selection will lead to a data leakage from train to test. Multiple random split \citep{gorman-bedrick-2019-need} may be a fairer approach, but this will make comparison of models more difficult. To address these problems, \citet{sogaard-etal-2021-need} advocates for the utilization of a biased or adversarial split besides the standard split, aiming to reduce the deviation between the test set and real-world data. We adopted this suggestion and developed an unbiased standard test set along with two biased challenge test sets, as detailed in Section~\ref{sec:Method}.

\subsection{Semantic Corpora with Long Texts}

Few corpora focus on the semantics of long texts, primarily because of difficult annotations and constraints in meaning representation itself (For instance, AMR was initially designed for single sentences). \citet{ogorman-etal-2018-amr} addressed this by manually annotating coreference, implicit roles, and bridging relations to create the multi-sentence AMR corpus. Other annotated corpora address discourse structure and rhetorical structure \cite{pdtb}, but ignore sentence semantics. As mentioned in Section~\ref{subsec:drs}, DRS is naturally designed for discourse, eliminating the need for additional annotation rules when annotating the meaning of long texts. Therefore, our annotation is more straightforward, as introduced in Section~\ref{sec:Method}.

\subsection{Compositional Generalization}

Several studies have demonstrated that neural models tend to memorize patterns observed during training, struggling to generalize effectively to unfamiliar patterns \citep{pmlr-v80-lake18a, Furrer-2020-com}. The combinationality in language significantly exacerbates this struggle. To assess this, tasks and datasets like the SCAN \citep{Lake2017GeneralizationWS} and the COGS \citep{kim-linzen-2020-cogs} have been developed. \citet{kim-linzen-2020-cogs} pointed out despite excellent standard test performances, their models reveal gaps in compositional generalization ability. This kind of gap led to our creation of the second challenge test set in Section~\ref{sec:Method} and experiments in Section~\ref{sec:eval}.


\section{Improving Semantic Evaluation\label{sec:Method}}

In this section we outline the methods to create better test sets.
Besides the standard test set created with a different data split, we also show how we built additional challenge test sets. The resulting data set will be released as PMB 5.0.0\footnote{The release is available at \url{https://pmb.let.rug.nl/releases/}}.




\subsection{Splitting Data Systematically}
As mentioned in Section~\ref{sec: introduction}, the random split method employed by the PMB requires improvement. We have devised a strategy that reduces overlap between training and standard development/test sets, without introducing additional biases.

Our data split strategy involves two rounds of sorting. First, documents are sorted by character length. Afterward, the ordered collections are divided into groups of ten documents, which are then re-sorted based on their internal edit distances. The first sorting aims to maintain a consistent length distribution across the training, development, and test sets, while also ensuring some degree of uniformity in their semantic distribution. This is crucial to minimize bias introduced in the standard test data. The second sorting is particularly designed to create a certain degree of separation between the datasets, aiming at decreasing the word overlap. We allocate the first eight documents to the training set, and the remaining two are randomly distributed between the development and test sets. In Section~\ref{sec:eval}, our experiments and analysis prove that the systematic split reduces the overlap between the training and development/test sets.

The distributions of gold data under the systematic split are shown in Table \ref{table: data split}. For English, we adopt an 8:1:1 split ratio, while for the other three languages, we use a 4:3:3 ratio to ensure the test data is sufficient.

\begin{table*}[hbtp]
\centering
\resizebox{0.8\textwidth}{!}{
\begin{tabular}{lrrrrr}
\toprule
& \multicolumn{1}{c}{\textbf{Train}} & \multicolumn{1}{c}{\textbf{Dev}} & \multicolumn{1}{c}{\textbf{Standard Test}} & \multicolumn{1}{c}{\textbf{Long Test}} & \multicolumn{1}{c}{\textbf{Compositional Test}} \\
\midrule
\textbf{English (EN)} & 9,057 (5.64) & 1,132 (5.38) & 1,132 (5.15) & 138 (60.78) & 1,148 (6.48) \\
\textbf{German (DE)} & 1,206 (5.06) & 900 (4.79) & 900 (4.87) & --- & --- \\
\textbf{Dutch (NL)} & 586 (5.62) & 435 (5.09) & 435 (5.08) & --- & --- \\
\textbf{Italian (IT)} & 745 (4.73) & 555 (4.52) & 555 (4.53) & --- & --- \\
\bottomrule
\end{tabular}
}
\caption{Distribution of train, development, and test sets in PMB 5.0.0 using the systematic split, together with two challenge sets. The average sentence length of each set are provided in brackets.}
\label{table: data split}
\end{table*}


\subsection{Creating Challenge Sets}
We create two challenge sets for English: one focusing on long texts and another dedicated to compositional recombination by CCG.

\begin{figure}[hbtp]
    \centering
    \includegraphics[scale=0.59]{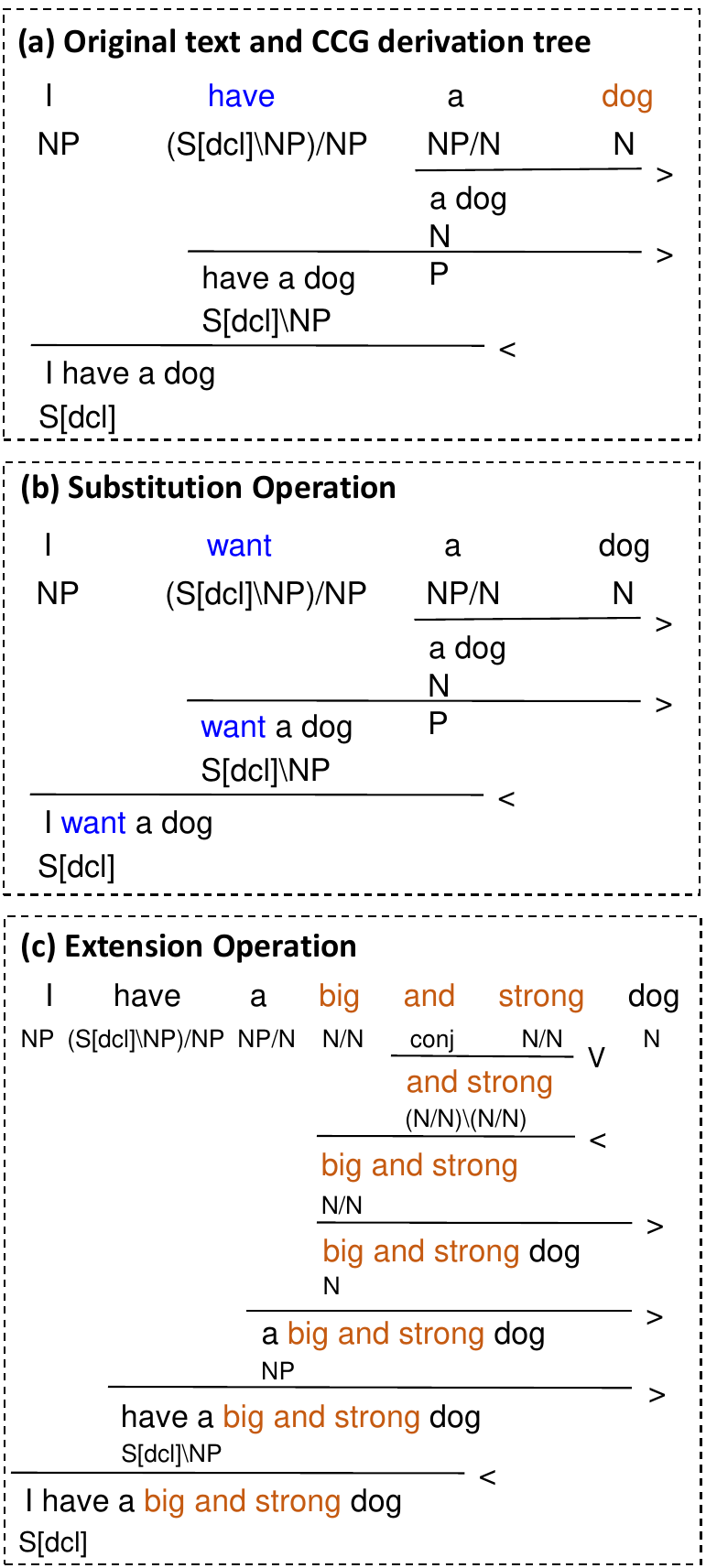} \vspace*{-20pt}
    \caption{Two recombination operations performed on the CCG derivation tree of example sentence ``\emph{I have a dog}'': (b) substitution (c) extension. We retained only the CCG categories and their corresponding words/phrases, excluding other semantic information.}
    \label{fig: ccg recombine}
\end{figure}

\vspace{-5pt}
\subsubsection{Long-Text Challenge Set}

Given that the gold data in the PMB predominantly consists of short sentences, with an average sentence length ranging between five and six words, it constrains our evaluation of the model's capability with long texts. In response, we select silver documents that notably exceed this average length for manual annotation, and change these into gold by correcting discourse structure, rhetorical relations, ellipsis, and inter-sentential pronouns (see Appendix~\hyperref[appendix: casestudy]{A.2} for an example).
Our long-text set includes 138 data samples with an average text length of 61 words, roughly ten times longer than the standard test set. The average lengths of train, development and test sets are shown  in Table \ref{table: data split}. 


\vspace{-5pt}

\subsubsection{Compositional Challenge Set}

As introduced in section \ref{sec: background},  the final layer of the PMB produces the CCG derivation tree that is enriched with syntactic and semantic information, which is subsequently passed to the boxer to produce DRS. Therefore, recombining the gold CCG tree with other trees can yield distinct CCG trees, with associated text and DRS. In contrast to the creation of the long-text set, the quality of the DRS produced by this method closely approximates the gold standard, which greatly reduces the need for further manual annotation.

The original CCG derivation tree contains the compositional categories of words and phrases in a sentence, as shown in Figure \ref{fig: ccg recombine} (a). We introduce two recombination operations: substitution and extension, shown in Figure \ref{fig: ccg recombine} (c) and (d). In the substitution operation, the leaves or subtrees within a CCG derivation tree are replaced by counterparts from other different trees, provided they share the same CCG category. For instance, the word \emph{have} swaps with \emph{want}, as highlighted in blue. The extension operation takes a singular leaf from the tree and develops it into a larger subtree. As shown in Figure \ref{fig: ccg recombine} (c), \emph{dog} with the $N$ category is extended to a subtree rooted at $N$, resulting in the phrase \emph{big and strong dog}. The pseudo-code detailing these two operations is provided in Appendix~\hyperref[appendix: ccg]{A.1}.

However, this method will generate many semantically abnormal sentences though they adhere strictly to syntactic structure. In this case, we use masked language models to estimate sentence pseudo-log-likelihood (PLL) scores \citep{salazar-etal-2020-masked, kauf2023better}. In practice, BERT \citep{devlin2018pretraining} is utilized as the scoring model, with a manually determined threshold. Specifically, the threshold is adjusted to eliminate 95\% of the generated sentences, retaining only the top 5\% that are highly deemed semantically correct.

Using this approach, we recombine the CCG trees of training samples and choose from the generated data, with the details presented in Table \ref{table: data split}. Table \ref{table: substitution} and \ref{table: extension} show some example texts produced through substitution and extension operations. Beyond individual operations, we also conduct multiple iterations on a sentence. The symbol $\times$ indicates the number of times an operation is applied to the same sentence. 

\begin{table*}[htbp]
\centering
\customfont
\begin{tabular}{c|c|ll}
\toprule
\textbf{Category} & \textbf{Operation} & \textbf{Training Set} & \textbf{Compositional Set} \\ \midrule
Noun & N$\Rightarrow$N  & Bill was killed by an \textcolor{blue}{intruder}. & Bill was killed by an \textcolor{blue}{Irishman}. \\
Pronoun & NP$\Rightarrow$NP & \textcolor{blue}{My} bag is very heavy. & \textcolor{blue}{His} bag is very heavy. \\
Verb & (S\textbackslash{}NP)/NP$\Rightarrow$(S\textbackslash{}NP)/NP & The police are \textcolor{blue}{following} us. & The police are \textcolor{blue}{visiting} us. \\
Adjective & S\textbackslash{}NP$\Rightarrow$S\textbackslash{}NP & My tie is \textcolor{blue}{orange}. & My tie is \textcolor{blue}{wet}. \\
Adverb & (S\textbackslash{}NP)/(S\textbackslash{}NP)$\Rightarrow$(S\textbackslash{}NP)/(S\textbackslash{}NP) & The rent is \textcolor{blue}{very} high. & The rent is \textcolor{blue}{extremely} high. \\
Preposition & PP/NP$\Rightarrow$PP/NP & The boy bowed \textcolor{blue}{to} me. & The boy bowed \textcolor{blue}{behind} me. \\
Determiners & NP/N$\Rightarrow$NP/N & \textcolor{blue}{The} answer is clear. & \textcolor{blue}{Neither} answer is clear. \\
Modal & (S\textbackslash{}NP)/(S\textbackslash{}NP)$\Rightarrow$(S\textbackslash{}NP)/(S\textbackslash{}NP) & It \textcolor{blue}{will} be scary. & It \textcolor{blue}{should} be scary. \\ \hline
Substitution$\times$2 & \begin{tabular}[c]{@{}c@{}}N$\Rightarrow$N\\ + (S\textbackslash{}NP)/NP$\Rightarrow$(S\textbackslash{}NP)/NP\end{tabular} & \textcolor{blue}{Russia} \textcolor{blue}{fears} the system. & \textcolor{blue}{Cuba} \textcolor{blue}{replaced} the system. \\ \hline
Substitution$\times$3 & \begin{tabular}[c]{@{}c@{}}NP$\Rightarrow$NP\\ + PP/NP$\Rightarrow$PP/NP\\ + S\textbackslash{}NP$\Rightarrow$S\textbackslash{}NP\end{tabular} & \begin{tabular}[c]{@{}l@{}}\textcolor{blue}{I} took \textcolor{blue}{the} elevator to the\\ \textcolor{blue}{fourth} floor.\end{tabular} & \begin{tabular}[c]{@{}l@{}}\textcolor{blue}{They} took \textcolor{blue}{another} elevator to\\ the \textcolor{blue}{last} floor.\end{tabular} \\ \bottomrule
\end{tabular}
\caption{Examples of substitution operations with CCG categories and operations. Note the table only shows the most common combinations for both two-fold (substitution $\times$ 2) and three-fold (substitution $\times$ 3) iterations. The color blue indicates the operation depicted in Figure \ref{fig: ccg recombine} (b).}
\label{table: substitution}
\end{table*}

\begin{table*}[htbp]
\customfont
\centering
\begin{tabular}{c|ll}
\toprule
\textbf{Category}                         & \textbf{Training Set}                          & \textbf{Compositional Set}                                                                                                                          \\ \midrule
Noun                             & My \textcolor{brown}{brother} is rich.                   & \begin{tabular}[c]{@{}l@{}}My \textcolor{brown}{bad} brother is rich.\\ My brother \textcolor{brown}{who is speaking English} is rich.\end{tabular}                              \\
Verb                             & Coffee will be \textcolor{brown}{served} after the meal. & \begin{tabular}[c]{@{}l@{}}Coffee will be \textcolor{brown}{secretly} served after the meal.\\ Coffee will be served \textcolor{brown}{by Elizabeth} after the meal.\end{tabular} \\
Adjective                        & Tom was \textcolor{brown}{thoughtful}.                   & \begin{tabular}[c]{@{}l@{}}Tom was \textcolor{brown}{very thoughtful}.\\ Tom was thoughtful \textcolor{brown}{and innocent}.\end{tabular}                                        \\ \hline
\multicolumn{1}{l|}{Extension$\times$2} & \textcolor{brown}{Tom} is \textcolor{brown}{courteous}.                     & \begin{tabular}[c]{@{}l@{}}Tom \textcolor{brown}{himself} is \textcolor{brown}{more} courteous.\\ Tom \textcolor{brown}{who did it} is  courteous.\end{tabular}                              \\ \hline
\multicolumn{1}{l|}{Extension$\times$3} & There are \textcolor{brown}{thirty} \textcolor{brown}{names} on the \textcolor{brown}{list}.   & \begin{tabular}[c]{@{}l@{}}There are \textcolor{brown}{about} thirty \textcolor{brown}{new} names on the \textcolor{brown}{short} list.\\ There are \textcolor{brown}{over} thirty \textcolor{brown}{other} names by \textcolor{brown}{Berlioz} on the list.\end{tabular} \\ \bottomrule
\end{tabular}
\caption{Examples of extension operations. We have excluded the operations of CCG categories due to the vast number of extension variations, which are nearly impossible to cover comprehensively. Instead, we present the most prevalent extension types for each category. The color orange indicates the operation depicted in Figure \ref{fig: ccg recombine} (c).}
\label{table: extension}
\end{table*}

\section{Experiments and Analysis \label{sec:eval}}
This section offers an introduction to the selected seq2seq models, experimental settings, results and analysis for the text-to-DRS parsing and DRS-to-text generation. 

\subsection{Model Selection\label{subsec: lm}}

The current approach to semantic parsing and text generation with DRS mainly involves fine-tuning a pre-trained language model. Our initial experiment employs a model based on BERT embeddings and LSTM architecture, following the methodology of  \citet{van-noord-etal-2020-character}. Then we utilize T5 and BART, two pre-trained transformer-based models. Specifically, we choose their multilingual variants: mT5 \citep{xue-etal-2021-mt5}, byT5 \citep{xue-etal-2022-byt5}, mBART \citep{liu-etal-2020-multilingual-denoising}, and DRS-MLM 
\citep{wang-etal-2023-pre} which is pre-trained on DRS data using the mBART architecture. In the case of DRS-MLM, for it is initially pre-trained on a train set under random split, we re-pre-train it using the train set based on our systematic split. To maintain consistent model sizes, we selected the large version across all models.

\subsection{Evaluation Metrics}

The evaluation process for Text-to-DRS parsing consists of two primary phases \citep{poelman-etal-2022-transparent}. Firstly, the generated DRSs and gold standard DRSs are transformed into Penman notation \citep{kasper-1989-flexible}. Subsequently, we utilize SMATCH \citep{cai-knight-2013-smatch}, an evaluation tool for AMR parsing, to calculate the match between the output and the gold standard by quantifying the overlap of triples. 
Evaluation of the generation task is conducted using BLEU \citep{Kishore-2002-BLUE}, METEOR \citep{lavie-agarwal-2007-meteor}, and COMET
\citep{rei-etal-2020-comet}.

\subsection{Experiment Settings}
We carried out three primary experiments. (1) We fine-tuned the selected language models for four languages: EN, DE, NL, and IT, and evaluated them using the standard test set. Following the training configurations set by \citet{rik-2018-DRS, poelman-etal-2022-transparent, wang-etal-2023-pre}, we trained the models on gold and silver data for EN, and trained on gold, silver, and bronze data for DE, NL, and IT. This was subsequently followed by a fine-tuning phase exclusively on gold data; (2) We calculated and compared the word overlap rate of the train sets and test sets under systematic and random split. Then, we showed the performance of the two top-performing models from the first experiments under these two splits. To ensure the assessment was solely influenced by the data split, we only tested on the English (only English has sufficient gold data) and fine-tuned exclusively on the gold data, and (3) We tested all fine-tuned models in the first experiments on the long-text set and compositional set. We divided the compositional set into two subsets: substitution and extension, to assess the difficulty produced by these two operations. 

For all experiments and models, uniform hyperparameters were employed, and the presented results are the average scores derived from three parallel experiments.\footnote{We provide the most recent experimental results for all test sets, available at  \url{https://pmb.let.rug.nl/models.php}.}

\begin{table*}[htbp]
\footnotesize
\centering
\setlength{\tabcolsep}{10pt}
\begin{tabular}{c|cc|cc|cc|cc}
\toprule
 & \multicolumn{2}{c|}{\textbf{English}} & \multicolumn{2}{c|}{\textbf{German}} & \multicolumn{2}{c|}{\textbf{Dutch}} & \multicolumn{2}{c}{\textbf{Italian}} \\
\textbf{Parser} & F1 & ERR & F1 & ERR & F1 & ERR & F1 & ERR \\ \midrule
LSTM & 78.6 & 8.4 & 80.2 & 4.0 & 74.4 & 8.5 & 79.6 & 5.0 \\
mT5 & 88.8 & 2.8 & 86.7 & 1.9 & 47.0 & 16.0 & 82.0 & 2.8 \\
byT5 & 91.4 & 2.1 & \textbf{88.0} & \textbf{0.7} & 79.8 & 5.0 & \textbf{87.2} & \textbf{0.7} \\
mBART & 89.1 & 2.3 & 86.1 & 1.8 & 64.5 & 3.4 & 86.2 & 1.8 \\
DRS-MLM & \textbf{91.5} & \textbf{1.5} & 87.1 & 2.1 & \textbf{85.5} & \textbf{2.0} & \textbf{87.2} & 0.9 \\ 
\bottomrule
\end{tabular}
\caption{Evaluation results for neural text-to-DRS parsing on the standard test sets of four languages. Note: ERR is the ill-formed rate (\%) of generated DRSs that fail to transform into a graph structure.}
\label{table: parsing result}
\end{table*}

\begin{table*}[!ht]
\centering
\footnotesize  
\begin{tabular}{c|ccc|ccc|ccc|ccc}
\toprule
 & \multicolumn{3}{c|}{\textbf{English}} & \multicolumn{3}{c|}{\textbf{German}} & \multicolumn{3}{c|}{\textbf{Dutch}} & \multicolumn{3}{c}{\textbf{Italian}} \\ 
     \textbf{Generator}                  & B       & M     & C    & B       & M       & C   & B       & M       & C   & B       & M       & C   \\ \midrule
LSTM                   & 33.8     & 32.4  & 72.5 & 24.9   & 25.4   & 67.1 &  19.0   & 21.6    & 63.2  &  28.2  &  24.7   & 72.2 \\
mT5                    & 69.9   & 53.4   & 92.8 & 47.8   & 37.5   & 84.8 &  11.3 &  15.2  & 63.6 &  48.8  &  36.3  & 86.0\\
byT5                   & \textbf{71.9}   & \textbf{54.9}   & \textbf{93.0} & \textbf{50.9} & \textbf{39.1}   & \textbf{85.2} &  41.8 & 34.2   & 82.1 & \textbf{53.2}  &  \textbf{38.5}    &  \textbf{87.5} \\
mBART                  & 51.8   & 43.5   & 88.1 & 40.8 & 33.4   & 79.9 & 38.1 &  32.0    & 80.6 &  45.8   &  34.5   & 84.7 \\
DRS-MLM                & 67.5   & 52.4   &  92.2   & 47.6   & 36.6   & 84.4 & \textbf{49.4}   & \textbf{37.5}   & \textbf{86.0}    & 46.3   & 34.2   &  86.3   \\ 
\bottomrule
\end{tabular}
\caption{Evaluation results for neural DRS-to-text generation on the standard test sets of four languages. Note: B = BLEU; M = METEOR; C = COMET.}
\label{table: generation result}
\end{table*}

\subsubsection{Standard Test}
Table \ref{table: parsing result} shows the results of the text-to-DRS parsing task. Across the four languages, both byT5 and DRS-MLM models stood out, with byT5 attaining 88.0 in German, slightly surpassing DRS-MLM's 87.1, and both models achieving the same F1 of 87.2 in Italian. However, in English and Dutch, DRS-MLM takes the lead with F1 91.5 and 85.5 respectively. mT5 and mBART closely follow, but their performance in Dutch is significantly weaker, possibly due to the limited Dutch data in their pre-training corpus.

Table \ref{table: generation result} shows the results of DRS-to-text generation. ByT5 surpasses other models in all languages except for Dutch. Particularly in English, ByT5 achieves top scores with 71.9, 54.9, and 93.0 in three metrics, respectively. However, for the Dutch, DRS-MLM remains the superior model across these three metrics.

The standout performance of byT5 and DRS-MLM can be attributed to byte-level tokenization and specific pre-training, respectively. Unlike other tokenization methods, like Byte Pair Encoding (BPE, \citealp{sennrich-etal-2016-neural}), byT5's byte-level tokenization, which can be seen as character-level within our four target languages, results in a smaller dictionary and has the ability to handle unseen words. 
DRS-MLM employs several pre-training tasks on the PMB data, making the model better suited for the DRS data format. This advantage is most obvious when dealing with Dutch, which has the least training data among the four languages.

\subsubsection{Systematic Split vs. Random Split}
\begin{figure}[htbp]
    \includegraphics[scale=0.250]{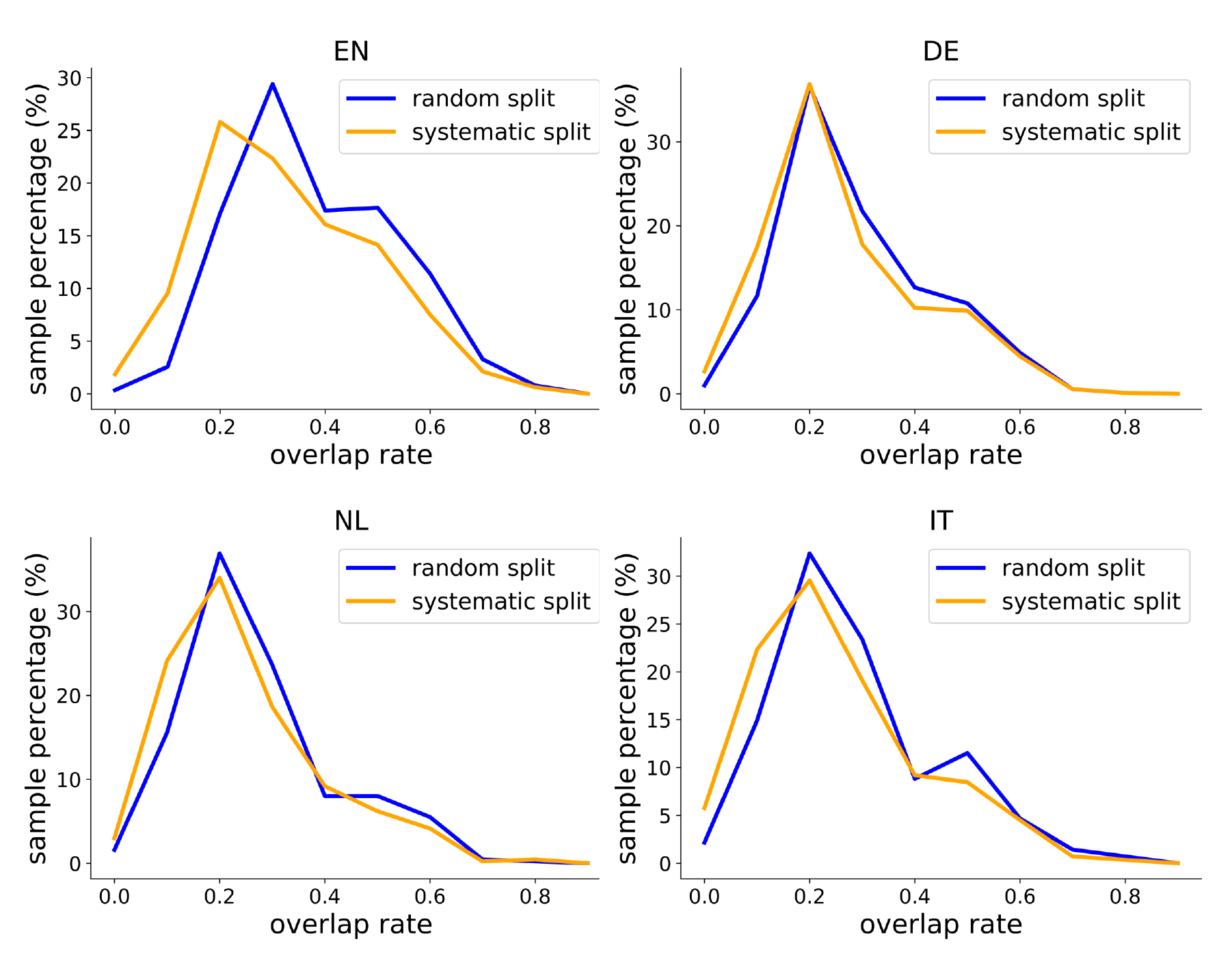}
    \caption{Distribution of word overlap rates between train and test sets in EN, DE, NL, IT. Lower overlap rates signify fewer words occurring in both train and test sets.}
    \label{fig: overlap}
\end{figure}

Figure \ref{fig: overlap} displays the distribution of word overlap rates between train and development/test sets under random and systematic split. The word overlap rate, defined in Equation \ref{equ: overlap}, measures the word-level sentence similarity. According to the figure, the systematic word overlap distribution is further to the left than the random split, indicating that it has less overlap. And as outlined in Section~\ref{sec:Method}, the systematic split does not simply reduce overlap by indiscriminately adding bias. It also guarantees that each set has a consistent length distribution, which can also be viewed as a semantic distribution to a certain extent. Therefore, in the case of PMB, a systematic split is a more effective method for dividing the dataset compared to the random split.

\vspace{-10pt}
\begin{equation} 
\centering
    \label{equ: overlap}
    \mathrm{overlap} = \frac{\mathrm{sentence1} \cap \mathrm{sentence2}}{\mathrm{sentence1} \cup \mathrm{sentence2}}
\end{equation}

We further proved the advantage through experiments. The parsing and generation results under these two splits are shown in Table \ref{table: systematic-parsing} and \ref{table: systematic-generation}. The model's performance on the random split exceeds that on the systematic split for both tasks, suggesting the systematic approach presents more rigorous challenges.

\begin{figure}[htbp]
    \includegraphics[scale=0.250]{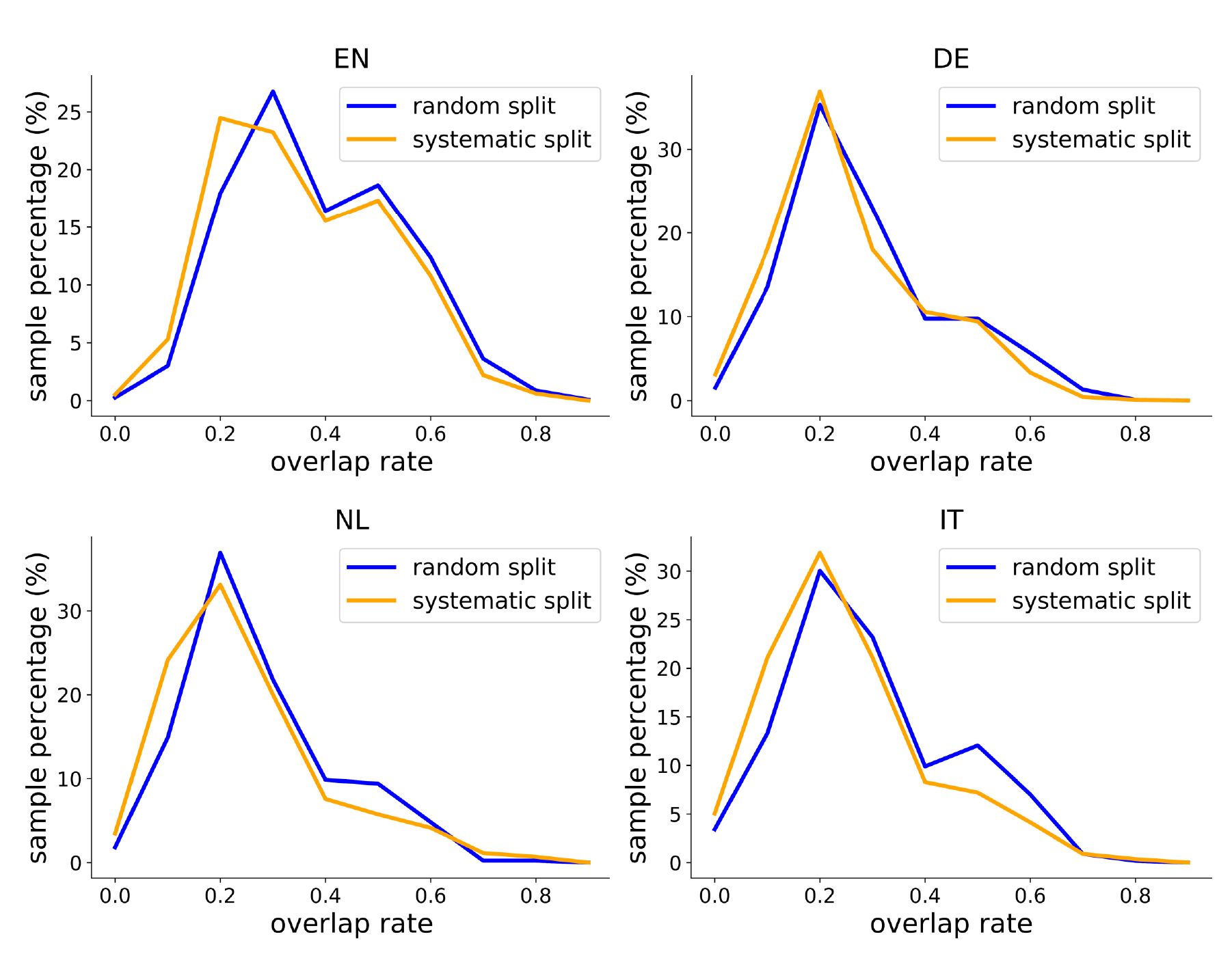}
    \caption{Distribution of word overlap rates between train and development sets in EN, DE, NL, IT.}
    \label{fig: overlap_dev}
\end{figure}

\begin{table}[htbp]
\centering
\resizebox{\columnwidth}{!}{
\begin{tabular}{c|cc|cc}
\toprule
 & \multicolumn{2}{c|}{\textbf{Random split}} & \multicolumn{2}{c}{\textbf{Systematic split}} \\
\textbf{Parser}                      & F1   & ERR   & F1   & ERR   \\ 
\midrule
byT5                    & 87.1 & 5.0   & \textbf{83.5} & \textbf{6.0}     \\
DRS-MLM                 & 88.9 & 1.9   & \textbf{87.3} & \textbf{4.1}   \\ 
\bottomrule
\end{tabular}
}
\caption{Results of parsing under random and systematic split. Lower scores are marked.}
\label{table: systematic-parsing}
\end{table}

\begin{table}[htbp]
\centering
\resizebox{\columnwidth}{!}{
\begin{tabular}{c|ccc|ccc}
\toprule
 & \multicolumn{3}{c|}{\textbf{Random split}} & \multicolumn{3}{c}{\textbf{Systematic split}} \\ 
\textbf{Generator}                           & B          & M          & C       & B           & M          & C          \\ 
\midrule
byT5                       & 66.1       & 52.2       & 91.7    & \textbf{64.7}        &  \textbf{51.0}      & \textbf{89.0}       \\
DRS-MLM                    & 65.8       & 51.4       & 91.7    & \textbf{60.2}        &  \textbf{48.4}      & \textbf{87.9}      \\ 
\bottomrule
\end{tabular}
}
\caption{Results of generation under random and systematic split.}
\label{table: systematic-generation}
\end{table}

\subsubsection{Challenge Test Sets}

The results of the models on the challenge test sets are shown in Tables \ref{table:challenge-parsing} and \ref{table:challenge-generation}.
The performance on the long-text test set is significantly inferior, marked by a high incidence of ill-formed outputs\footnote{SMATCH employs a hill-climbing technique to identify the optimal match, which may introduce inaccuracies when evaluating the output of the model for long texts \cite{opitz-frank-2022-better}. In this case, the results for long texts should be considered as reference only.}. The most pronounced drop is observed in ByT5, which shows a reduction of 86\% compared to the standard test set. In the generation task, although truncation does not hugely impact on evaluation, the models still grapple with long sequences, reflecting decreases of at least 29.9, 11.9, and 16.2 across three metrics. Notably, neural models struggle with the long set, primarily because their tokenization significantly amplifies both input and output lengths. For example, while the average sentence lengths in the long set stand at 61 for text and 253 for DRS, these numebrs increase to 98 and 503 after BPE tokenization (mT5, mBART, and DRS-MLM) and even further to 410 and 1370 with character-level tokenization (ByT5). Obviously, these models can not handle such long sequences as effectively as the short sequences in the standard test.


For the compositional challenge set, it's crucial to note that all semantic components in the test sets were also in the training. Therefore, we expect near-perfect scores from the models. They perform well on the \emph{compositional-substitution} set, showcasing their ability to learn and apply word meanings in known sentence structures. Among these models, byT5 performs the best with 93.1 F1 in parsing, while mT5 and DRS-MLM show similarly strong performance in generation. When testing on the \emph{compositional-extension} set, the performance of the models dropped by around ten points in both tasks. Most parsing or generation errors were in the newly added parts in the texts, likely due to the introduction of more intricate sentence structures, especially compound predicate adjectives and attributive clauses, as shown in the examples in Table~\ref{table: extension}.
The most frequent errors of the models are provided with examples in Appendix~\hyperref[appendix: casestudy]{A.2}.


\begin{table}[htbp]
\centering 
\resizebox{\columnwidth}{!}{
\begin{tabular}{c|cc|cc|cc}
\toprule
 & \multicolumn{2}{c|}{\textbf{en-long}} & \multicolumn{2}{c|}{\textbf{en-substitution}} & \multicolumn{2}{c}{\textbf{en-extension}} \\

\textbf{Parser} & F1 & ERR & F1 & ERR & F1 & ERR \\
\midrule
LSTM & \textbf{43.7} & \textbf{19.2} & 90.8 & 2.8 & 82.7 & \textbf{3.5} \\
mT5 & 38.8 & 34.6 & 88.9 & 2.9 & 80.3 & 8.9 \\
byT5 & 5.5 & 65.4 & \textbf{93.1} & \textbf{0.5} & \textbf{84.8} & 5.0 \\
mBART & 22.0 & 53.8 & 89.7 & 1.4 & 80.4 & 7.6 \\
DRS-MLM & 20.0 & 57.7 & 90.3 & 2.8 & 81.1 & 7.7 \\
\bottomrule
\end{tabular}
}
\caption{Evaluation results for text-to-DRS parsing on the challenge test sets.}
\label{table:challenge-parsing}
\end{table}

\begin{table}[htbp]
\centering
\setlength{\tabcolsep}{4pt}
\resizebox{\columnwidth}{!}{
\begin{tabular}{c|ccc|ccc|ccc}
\toprule
 & \multicolumn{3}{c|}{\textbf{en-long}} & \multicolumn{3}{c|}{\textbf{en-substitution}} & \multicolumn{3}{c}{\textbf{en-extension}} \\ 
\textbf{Generator}                       & B        & M        &  C     & B          & M           & C          &   B         & M         & C         \\ 
\midrule
LSTM                   & 5.48     & 14.6     & 40.3  &  58.7      &  43.6       &  82.1      &  49.1       & 41.3      &  77.6     \\
mT5                    & 31.4     & 40.3     & \textbf{76.6}   &  75.2      &  \textbf{55.6}       &  \textbf{92.7}     &  67.3       & 52.9      &   \textbf{90.0}    \\
byT5                   & 14.1     & 28.3     & 59.3   &  75.7      &  54.7       &  92.5     &  66.7       & 53.0      & 89.8          \\
mBART                  & 15.7     & 28.7     & 60.6   &  68.8      &  51.8       &  89.8     &  58.4       & 48.8      &  86.1    \\
DRS-MLM                & \textbf{32.6}     & \textbf{40.5}     & 75.4   &  \textbf{76.0}      &  54.9       &  92.5     &  \textbf{69.4}       & \textbf{53.2}      &   \textbf{90.0}   \\ 
\bottomrule
\end{tabular}
}
\caption{Evaluation results for DRS-to-text generation on the challenge test sets.}
\label{table:challenge-generation}
\end{table}



\vspace{-5pt}

\section{Conclusion}


Past performance of neural semantic parsers and meaning-to-text generators have been slightly inflated (or at best, made the suggestion that these semantic computational tasks were close to being ``solved'') due to data leakage from training to test and non-representative test sets. At least, that is what our empirical study on the Parallel Meaning Bank showed. We created a more realistic assessment of performance by refining the data split and formulating challenge sets. A systematic split for the PMB yields a test set that is harder for semantic parsers and generators. The introduction of two further challenge sets, one with manually corrected longer documents and one with automatically derived compositional recombination using categorical grammar, are indeed way more challenging than the standard test set.  Hence, semantic parsing and text-to-meaning generation can not be considered ``solved'' yet. 


\section{References}
\vspace{-20pt}
\bibliographystyle{lrec-coling2024-natbib}
\bibliography{lrec-coling2024-example}

\bibliographystylelanguageresource{lrec-coling2024-natbib}

\appendix
\section*{A. Appendix}




\subsection*{Appendix A.1 Pseudo-code for CCG recombination \label{appendix: ccg}}
Both substitution and extension operations begin with a standard pre-processing step: subtree set construction. This extracts all subtrees from the dataset's CCG derivation trees (For consistency, we treat leaves as subtrees with only the root). Substitution operation primarily involves randomly selecting subtrees, and then deleting and substituting them. The replacement subtree is chosen from the list in the first step. Extension operation involves forming child mappings and producing subtrees according to the mappings.

\begin{algorithm}
\caption{Extract Subtrees from CCG Trees}
\begin{algorithmic}[1]
\State \textbf{Variables:}
\State $SubtreeList \gets$ empty list
\State $AllCCGTrees \gets$ CCG tree list
\\
\Function{ExtractSubs}{$node$, $currentPath$}
    \If{$node$ is null}
        \Return
    \EndIf
    
    \State Add $node$ to $currentPath$
    
    \If{$node.left$ and $node.right$ are null}
        \State Add $currentPath$ to $SubtreeList$
    \EndIf

    \State \Call{ExtractSubs}{$node.left$, $currentPath$}
    \State \Call{ExtractSubs}{$node.right$, $currentPath$}
\EndFunction
\\
\Function{SubtreesForTree}{$root$}
    \State \Call{ExtractSubs}{$root$, empty list}\\
    \Return $SubtreeList$
\EndFunction
\\
\Function{SubtreesForTrees}{$AllCCGTrees$}
    \For{each $tree$ in $AllCCGTrees$}
        \State \Call{SubtreesForTree}{$tree$}
    \EndFor \\
    \Return $SubtreeList$
\EndFunction

\end{algorithmic}
\end{algorithm}

\begin{algorithm}
\caption{Substitution Operation}
\begin{algorithmic}[1]
\State \textbf{Variables:}
\State $SubtreeList \gets$ list of subtrees
\\
\Function{GetParent}{tree, childNode}
    \For{each node \( n \) in tree}
        \If{\( n \).left = childNode or \( n \).right = childNode}
            \State \Return \( n \)
        \EndIf
    \EndFor
    \State \Return null
\EndFunction
\\
\Function{DeleteAndAdd}{tree, nodeToDelete}
    \State parent $\gets$ \Call{GetParent}{tree, nodeToDelete}
    \State newSubTree $\gets$ randomly select from $SubtreeList$ with same root of nodeToDelete
    
    \If{parent.left = nodeToDelete}
        \State parent.left $\gets$ newSubTree
    \ElsIf{parent.right = nodeToDelete}
        \State parent.right $\gets$ newSubTree
    \EndIf
\EndFunction
\\
\Function{Substitute}{tree}
    \State nodeToDelete $\gets$ randomly select a node from tree
    \State \Call{DeleteAndAdd}{tree, nodeToDelete}
\EndFunction

\end{algorithmic}
\end{algorithm}

\begin{algorithm}
\caption{Extension Operation}
\begin{algorithmic}[1]

\State \textbf{Variables:}
\State $Subtrees \gets$ list of subtrees
\State $ChildMap \gets$ dictionary of children
\\
\Function{Traverse}{node}
    \If{node is null}
        \State \Return
    \EndIf
    \If{node.left}
        \State $ChildMap[(node, node.left)] \gets$ node.right
    \EndIf
    \If{node.right}
        \State $ChildMap[(node, node.right)] \gets$ node.left
    \EndIf \\
    \Call{Traverse}{node.left}\\
    \Call{Traverse}{node.right}
\EndFunction
\\
\Function{CreateSubtree}{parent, left, right}
    \State parent.left = left
    \State parent.right = right
\EndFunction
\\
\Function{Extension}{tree}
    \State $leaf \gets$ \Call{RandomSelectLeaf}{tree}
    \If{left}
        \State $newSubRoot \gets$ \Call{CreateSubtree}{leaf, leaf, $ChildMap[(leaf, leaf)]$}
        \Comment{To extend the node from right}
    \Else
        \State $newSubRoot \gets$ \Call{CreateSubtree}{leaf, $ChildMap[(leaf, leaf)]$, leaf}
        \Comment{To extend the node from left}
    \EndIf
    \State choose the $newSubtree$ from $Subtrees$ according to $newSubRoot$
    \State replace $leaf$ with $newSubtree$    
\EndFunction

\end{algorithmic}
\end{algorithm}



\newpage

\subsection*{Appendix A.2 Case Study \label{appendix: casestudy}}

In this appendix, we present some wrong generations by byT5 model in the semantic parsing task. Additionally, the gold-standard text and DRS can also be seen as examples of the challenge sets.

\begin{table*}
\fontsize{6.5pt}{8pt}\selectfont 
\begin{tabular}{clll}
\toprule
Test set          & Gold Text& \multicolumn{1}{c}{Gold DRS}& \multicolumn{1}{c}{Generated}\\ \hline
Standard          & Mary called us.& \begin{tabular}[c]{@{}l@{}}female.n.02 Name "Mary"\\ call.v.03 Agent -1 Time +1 Co-Agent +2\\ time.n.08 TPR now\\ person.n.01 Sub speaker\end{tabular}& \begin{tabular}[c]{@{}l@{}}female.n.02 Name "Mary"\\ call.v.03 Agent -1 Time +1 Theme +2\\ time.n.08 TPR now\\ person.n.01 Sub speaker\end{tabular}\\
\multicolumn{1}{l}{}   &&&\\
Long Text         & \begin{tabular}[c]{@{}l@{}}Recent studies show\\ that children who do\\ not get enough sleep\\ tend to have some \\ emotional problems \\ as well as weight \\ gain later in life.  As \\ VOA's Melinda Smith\\ reports, the research\\ seems to blame the \\ parents.\end{tabular} & \begin{tabular}[c]{@{}l@{}}recent.a.02 AttributeOf +1\\ study.n.01\\ show.v.02 Proposition \textgreater{}1 Experiencer -1 Time +1\\ time.n.08 EQU now\\ CONTINUATION \textless{}0\\ child.n.01\\ tend.v.01 Agent -1 Time +1 Topic +2\\ time.n.08 EQU now\\ have.v.01 Pivot -3 Theme +3 Theme +7\\ emotional.a.03 AttributeOf +1\\ problem.n.01\\ entity.n.01 Sub -1 Sub +2\\ weight.n.01\\ gain.n.01 Theme -1\\ later.r.01 EQU -6\\ life.n.01\\ NEGATION \textless{}1\\ time.n.08 EQU now\\ get.v.01 Pivot -12 Time -1 Theme +2\\ enough.a.01 AttributeOf +1\\ sleep.n.01\\ CONTINUATION \textless{}3\\ agency.n.01 Name "VOA"\\ female.n.02 Name "Melinda Smith" PartOf -1\\ report.v.01 Agent -1 Time +1\\ time.n.08 EQU now\\ CONTINUATION \textless{}1\\ research.n.01\\ seem.v.01 Experiencer -1 Time +1 Stimulus +2\\ time.n.08 EQU now\\ blame.v.01 Agent -3 Theme +1\\ person.n.01 Role +1\\ parent.n.01\end{tabular} & \begin{tabular}[c]{@{}l@{}}recent.a.01 AttributeOf +1\\ study.n.04\\ show.v.04 Proposition \textgreater{}1 Experiencer -1 Time +1\\ time.n.08 EQU now\\ CONTINUATION \textless{}0\\ child.n.01\\ NEGATION \textless{}1\\ time.n.08 EQU now \\ get.v.01 Pivot -2 Time -1 Theme +2\\ enough.a.01 AttributeOf +1\\ sleep.n.01\\ tend.v.01 Agent -4\\ T\end{tabular} \\
\multicolumn{1}{l}{}   &&&\\
Substitution & \begin{tabular}[c]{@{}l@{}}Hungarian prisoners \\ broke out of jail.\end{tabular}& \begin{tabular}[c]{@{}l@{}}country.n.02 Name "Hungary"\\ person.n.01 Location -1 Role +1\\ prisoner.n.01\\ break\_out.v.03 Theme -2 Time +1 Source +2\\ time.n.08 TPR now\\ jail.n.01\end{tabular}& \begin{tabular}[c]{@{}l@{}}country.n.02 Name "Hungary"\\ person.n.01 Source -1 Role +1\\ prisoner.n.01\\ break\_out.v.01 Source -2 Time +1 Theme +2\\ time.n.08 TPR now\\ jail.n.01\end{tabular}\\
\multicolumn{1}{l}{}   &&&\\
Extension    & \begin{tabular}[c]{@{}l@{}}Mr. Smith who worked \\ on that project asked\\ Jane to marry him.\end{tabular}& \begin{tabular}[c]{@{}l@{}}mr.n.01\\ male.n.02 Name "Smith" Title -1\\ work.v.01 Agent -1 Time +1 Theme +2\\ time.n.08 TPR now\\ project.n.01\\ ask.v.02 Agent -4 Time +1 Recipient +2 Topic +3\\ time.n.08 TPR now\\ female.n.02 Name "Jane"\\ marry.v.01 Agent -1 Co-Agent +1\\ male.n.02 ANA -8\end{tabular}& \begin{tabular}[c]{@{}l@{}}mr.n.01\\ male.n.02 Name "Smith" Title -1\\ work.v.02 Agent -1 Time +1 Theme +2\\ time.n.08 TPR now\\ project.n.01\\ ask.v.02 Agent -4 Time +1 Patient +2 Result +3\\ time.n.08 TPR now\\ female.n.02 Name "Jane"\\ marry.v.01 Agent -1 Co-Agent +1\\ male.n.02 ANA -5\end{tabular}                                                   \\ \hline
\end{tabular}
\caption{Four examples in different test sets.}
\end{table*}

\end{document}